\documentclass[10pt,twocolumn,letterpaper]{article}

\usepackage[pagenumbers]{cvpr}  

\usepackage{capt-of}
\usepackage{xspace}
\usepackage{xcolor}
\usepackage{enumitem}
\usepackage{amsmath,amsthm,amssymb,amsfonts,dsfont,pifont,bm,bbm,mathrsfs,mathtools,nicefrac,extarrows,relsize}
\usepackage{algorithm,algpseudocode,listings}
\usepackage{booktabs,multirow,adjustbox,diagbox,threeparttable,tabularray,setspace}
\usepackage{CJK}

\definecolor{cvprblue}{rgb}{0.21,0.49,0.74}
\usepackage[pagebackref,breaklinks,colorlinks,citecolor=cvprblue,bookmarks=false]{hyperref}
\usepackage{wrapfig}
\usepackage[capitalize]{cleveref}  

\crefname{section}{Sec.}{Secs.}
\Crefname{section}{Section}{Sections}
\crefname{appendix}{App.}{Apps.}
\Crefname{appendix}{Appendix}{Appendices}
\crefname{table}{Tab.}{Tabs.}
\Crefname{table}{Table}{Tables}
\crefname{figure}{Fig.}{Figs.}
\Crefname{figure}{Figure}{Figures}
\crefname{equation}{Eq.}{Eqs.}
\Crefname{equation}{Equation}{Equations}
\crefname{theorem}{Thm.}{Thms.}
\Crefname{theorem}{Theorem}{Theorems}
\crefname{lemma}{Lem.}{Lems.}
\Crefname{lemma}{Lemma}{Lemmas}
\crefname{remark}{Rem.}{Rems.}
\Crefname{remark}{Remark}{Remarks}
\crefname{corollary}{Cor.}{Cors.}
\Crefname{corollary}{Corollary}{Corollaries}
\crefname{algorithm}{Alg.}{Algs.}
\Crefname{algorithm}{Algorithm}{Algorithms}
\definecolor{cellred}{RGB}{213, 123, 101}
\definecolor{cellgreen}{RGB}{0, 205, 0}
\definecolor{cellblue}{RGB}{54, 125, 189}
\definecolor{codegreen}{rgb}{0,0.6,0}
\definecolor{codegray}{rgb}{0.5,0.5,0.5}
\definecolor{codepurple}{rgb}{0.58,0,0.82}
\definecolor{backcolour}{rgb}{1.0,1.0,1.0}
\lstdefinestyle{mystyle}{
    backgroundcolor=\color{backcolour},
    commentstyle=\color{codegreen},
    keywordstyle=\color{magenta},
    numberstyle=\tiny\color{codegray},
    stringstyle=\color{codepurple},
    basicstyle=\ttfamily\scriptsize,
    breakatwhitespace=false,
    breaklines=true,
    captionpos=b,
    keepspaces=true,
    numbers=left,
    numbersep=5pt,
    showspaces=false,
    showstringspaces=false,
    showtabs=false,
    tabsize=2
}
\newcommand{\tocite}[1]{{\color{red} [TO CITE]}}

\title{\vspace{-1.3cm}Dynamic Typography: Bringing Text to Life via Video Diffusion Prior\vspace{-0.5cm}}

\author{Zichen Liu$^{*,1}$ \and
Yihao Meng$^{*,1}$ \and
Hao Ouyang$^{2}$ \and 
Yue Yu$^{1}$ \and
Bolin Zhao$^{1}$ \and
Daniel Cohen-Or$^{\dagger, 3}$ \and Huamin Qu$^{\dagger, 1}$ \and \hspace{0.9\linewidth}
\and $^{1}$ HKUST \and $^{2}$ Ant Group \and $^{3}$Tel-Aviv University
}

\begin{document}
\twocolumn[{
    \renewcommand\twocolumn[1][]{#1}
    \maketitle
    \begin{center}
      \vspace{-0.7cm}
      \includegraphics[width=\textwidth]{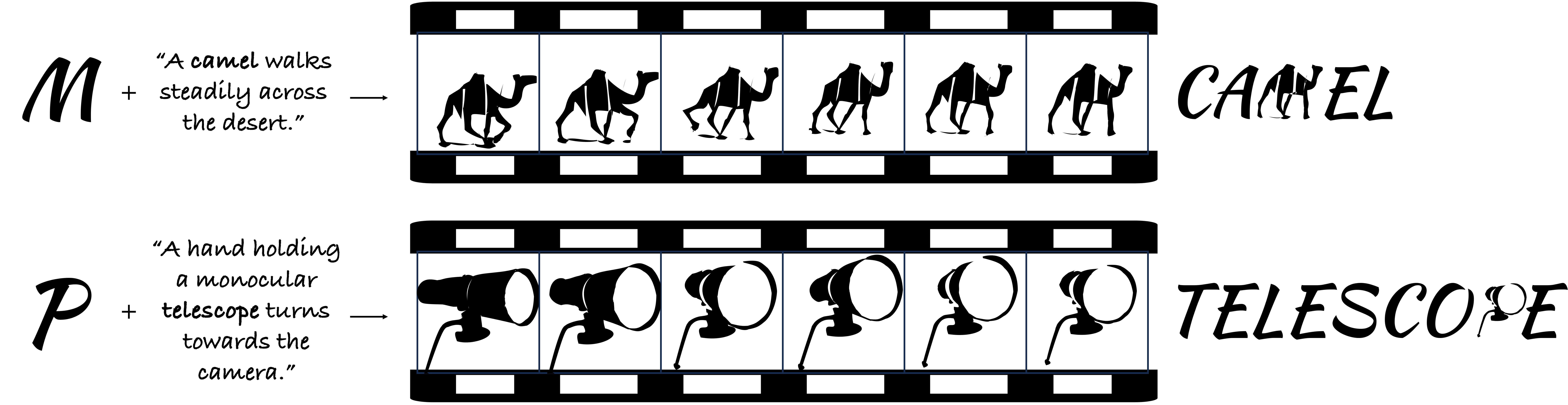}
      \vspace{-0.3cm}
      \captionsetup{type=figure}
      \caption{Given a letter and a text prompt that briefly describes the animation, our method automatically semantically reshapes a letter and animates it in vector format while maintaining legibility. Our approach allows for a variety of creative interpretations that can dynamically bring words to life. Our code is available at:
\textcolor{magenta}{\href{https://animate-your-word.github.io/demo/}{https://animate-your-word.github.io/demo/}}.}
      \label{fig:teaser}
    \end{center}
}]

\maketitle
\def\thefootnote{*}\footnotetext{Indicates Equal Contribution.}
\def\thefootnote{$\dagger$}\footnotetext{Indicates Corresponding Author.}
\begin{abstract}
Text animation serves as an expressive medium, transforming static communication into dynamic experiences by infusing words with motion to evoke emotions, emphasize meanings, and construct compelling narratives. Crafting animations that are semantically aware poses significant challenges, demanding expertise in graphic design and animation.
We present an automated text animation scheme, termed ``Dynamic Typography'', which combines two challenging tasks. It deforms letters to convey semantic meaning and infuses them with vibrant movements based on user prompts.
Our technique harnesses vector graphics and an end-to-end optimization-based framework. This framework employs neural displacement fields to convert letters into base shapes and applies per-frame motion, encouraging coherence with the intended textual concept. Perceptual loss regularization and shape preservation techniques are employed to maintain legibility and structural integrity throughout the animation process.
We demonstrate the generalizability of our approach across various text-to-video models and highlight the superiority of our end-to-end methodology over baseline methods, which might comprise separate tasks.
Through quantitative and qualitative evaluations, we demonstrate the effectiveness of our framework in generating coherent text animations that faithfully interpret user prompts while maintaining readability.
\end{abstract}
\section{Introduction}

Text animation is the art of bringing text to life through motion.
By animating text to convey emotion, emphasize meaning, and create a dynamic narrative, text animation transforms static messages into vivid, interactive experiences \cite{kTypoEngine, kTypoEmo}. 
The fusion of motion and text, not only captivates viewers, but also deepens the message's impact, making text animation prevail in movies, advertisements, website widgets, and online memes \cite{wakey_2023}. 

This paper introduces a specialized text animation scheme that focuses on animating individual letters within words. 
This animation is a compound task: The letters are deformed to embody their semantic meaning and then brought to life with vivid movements based on the user's prompt. 
We refer to it as ``Dynamic Typography''. 
For example, the letter ``M'' in ``CAMEL'' can be animated with the prompt ``A camel walks steadily across the desert'' in Fig. \ref{fig:teaser}. 
This animation scheme opens up a new dimension of textual animation that enriches the reading experience. 

However, crafting such detailed and prompt-aware animations is challenging. 
Traditional text animation methods demand considerable expertise in graphic design and animation \cite{kTypoEngine}, making them less accessible to non-experts.
Our methodology aims to automate the text animation process to make it more accessible and efficient. 
Following prior research in font generation and stylization \cite{learnedsvg, deepvecfont, wordAI}, we represent each input letter and every output frame as a vectorized, closed shape by a collection of B\'ezier curves. 
This vector representation is resolution-independent, keeping text clear at any scale and offering substantial editability, as users can easily adjust text appearance through control points.
However, this shift to vector graphics introduces unique challenges in text animation. 
Most current video generation methods \cite{I2V_LDM, videocomposer, videocrafter, dynamicrafter, animatediff} fall short in this new scenario as they are designed within the rasterized pixel-based scenario instead of vectorized shapes, and are hard to render readable text.
Although the most recent work, LiveSketch \cite{livesketch}, introduces an approach to animate arbitrary vectorized sketches, it struggles to preserve legibility and consistency throughout animation when the input becomes vectorized letters, causing visually unpleasant artifacts including flickering and distortion.

To address these challenges, we design an end-to-end optimization-based framework that utilizes two neural displacement fields, represented in coordinates-based MLP.
The first field deforms the original letter into the base shape, setting the stage for animation.
Subsequently, the second neural displacement field learns the per-frame motion applied to the base shape.
The two fields are jointly optimized using the score-distillation sampling (SDS) loss \cite{dreamfusion} to integrate motion priors from a pre-trained text-to-video model \cite{modelscope} to encourage the animation to align with the intended textual concept.
To preserve the legibility of the letter throughout the animation, we apply perceptual loss \cite{lpips} as a form of regularization on the base shape to maintain a perceptual resemblance to the original letter. 
Additionally, to preserve the overall structure and appearance during animation, we introduce a novel shape preservation regularization based on the triangulation \cite{mips} of the base shape, which forces the deformation between the consecutive frames to adhere to the principle of being conformal with respect to the base shape.
 
Our approach is designed to be data-efficient, eliminating the need for additional data collection or the fine-tuning of large-scale models. 
Furthermore, our method generalizes well to various text-to-video models, enabling the incorporation of upcoming developments in this area. 
We quantitatively and qualitatively tested our text animation generation method against various baseline approaches, using a broad spectrum of prompts.
The results demonstrate that the generated animation accurately and aesthetically interprets the input text prompt descriptions. 
Our method outperforms various baseline models in preserving legibility and prompt-video alignment.
Overall, our framework demonstrates its efficacy in producing coherent text animations from user prompts while maintaining the readability of the text, which is achieved by the key design of the learnable base shape and associated shape preservation regularization.

\section{Related Work}
\subsection{Static Text Stylization}
Text stylization enhances text aesthetics while maintaining readability, including artistic text style transfer and semantic typography. 
Artistic text style transfer migrates stylistic elements from source images onto text, typically using texture synthesis \cite{sketchPatch,Awesome_Typography} and GANs \cite{Multi_Content_GAN, SCFont, Intelligent_Typography,Typography_With_Decor}.
Semantic typography combines semantic understanding with visual representation, creating visual forms that convey meaning.
Notable works include Word-as-Image \cite{wordAI}, which uses Score Distillation Sampling \cite{dreamfusion} with diffusion prior \cite{Stable_diffusion} to create meaningful letter deformations, and DS-Fusion \cite{DS-fusion}, which employs latent diffusion to blend semantic-related styles into glyphs.

While these works produce only static images with limited semantic expression, our Dynamic Typography introduces motion to text, enhancing viewer engagement and aesthetic appeal \cite{dy_advantage}.

\subsection{Dynamic Text Animation}
Given animations' effectiveness in capturing attention \cite{animation}, research has explored dynamic text animations, particularly in dynamic style transfer and kinetic typography. 
Dynamic style transfer adapts visual style and motion from reference videos to text, with works like \cite{DynTypo} transferring animations from dynamic text videos, and \citet{shape-match} introducing a scale-aware Shape-Matching GAN for diverse styles.

Kinetic typography \cite{Kinetic_typography} integrates motion with text to enhance messages. 
While traditionally labor-intensive, recent works \cite{kinetic_engine, Kinetic_typography, Automatic_kinetic_typography, The_kinedit_system, emordle} aim to automate this process. 
For instance, Wakey-Wakey \cite{wakey_2023} uses a motion transfer model \cite{FOMM} to animate text using meme GIFs.

However, these approaches require specific driven videos and are limited to simple motion patterns. 
In contrast, our method generates arbitrary motion patterns using only text prompts as input.

\begin{figure*}[th!]
\centering
\includegraphics[width=1.0\textwidth]{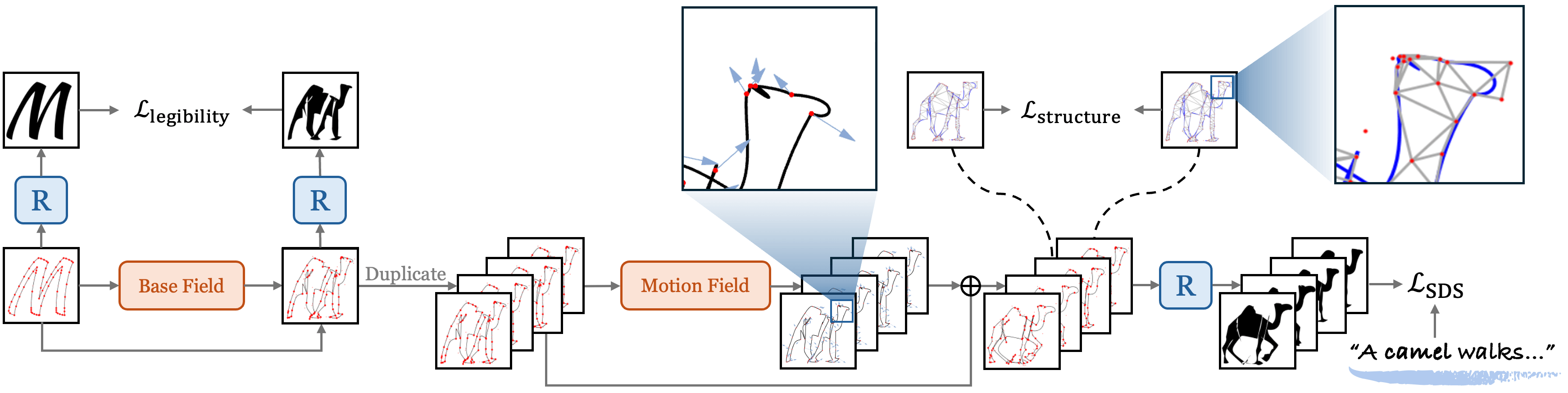}
\caption{An overview of the framework. Given a letter represented as a set of control points, the Base Field deforms it to the shared base shape, setting the stage to add per-frame displacement. Then the base shape is duplicated across $k$ frames, and the Motion Field predicts the displacement for each control point at each frame, infusing movement into the base shape. Each frame is rendered by the differentiable rasterizer $R$ and concatenated as the output video. The base and motion field are jointly optimized by the video prior from frozen pre-trained video foundation model using Score Distillation Sampling $\mathcal{L}_{\text{SDS}}$, under regularization on legibility $\mathcal{L}_{\text{legibility}}$ and structure preservation $\mathcal{L}_{\text{structure}}$.}
\label{model_overview}
\vspace{-0.3cm}
\end{figure*} 

\subsection{Text and Image-to-Video Generation}
Text-to-Video generation has advanced significantly with diffusion models. 
Several approaches extend Stable Diffusion (SD) \cite{Stable_diffusion} by incorporating temporal information, including AnimateDiff \cite{animatediff}, LVDM \cite{LVDM}, MagicVideo \cite{MagicVideo}, VideoCrafter \cite{videocrafter}, and ModelScope \cite{modelscope}. 
Image-to-Video generation methods like DynamiCrafter \cite{dynamicrafter}, Motion-I2V \cite{Motion-I2V}, Gen-2 \cite{gen-2}, Pika Labs \cite{pika}, and SVD \cite{svd} generate videos from images and prompts.

However, existing open-source models fail to maintain text readability during motion. 
Training models for text animation requires large text animation datasets, which are scarce. 
One recent work LiveSketch \cite{livesketch} animates arbitrary vectorized sketches without extensive training. 
This work leverages the motion prior from pre-trained text-to-video diffusion model using score distillation sampling \cite{dreamfusion} to guide the motion of input sketches.
However, when the input becomes vectorized letters, LiveSketch struggles to preserve legibility and consistency throughout the animation, leading to flickering and distortion artifacts that severely degrade video quality.
Our method, in contrast, generates consistent, prompt-aware text animations while preserving readability.
\section{Preliminary}

\subsection{Vector Representation and Fonts}
Modern font formats like TrueType \cite{truetype} and PostScript \cite{adobe1990postscript} utilize vector graphics to define glyph outlines.
These outlines are typically collections of B\'ezier or B-Spline curves, enabling flexible text rendering at any scale \cite{wordAI}. 
Our method outputs each animation frame in the same vector representation as our input. 

\begin{figure}[H]
\centering
\vspace{-0.6cm}
\includegraphics[width=0.65\columnwidth]{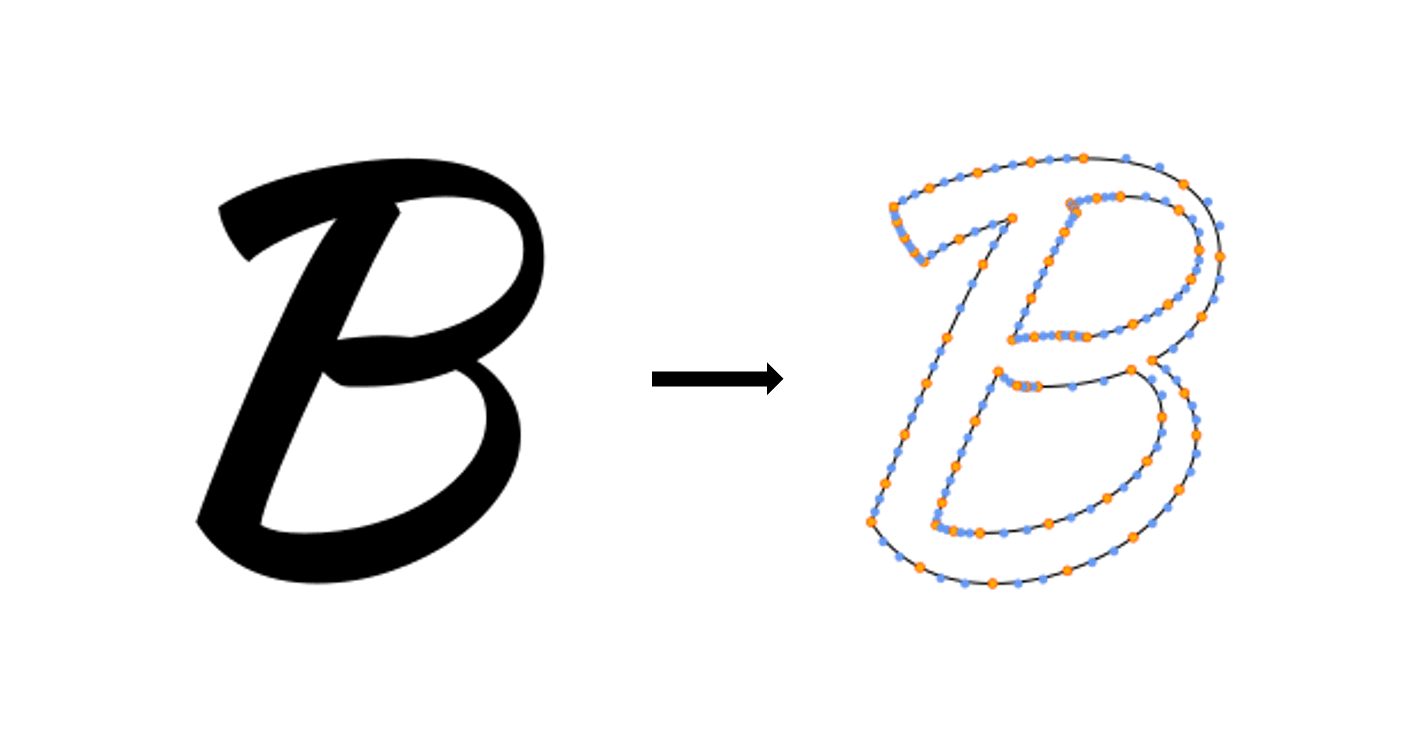}
\vspace{-0.8cm}
\caption{B\'ezier curves representation of letter ``B''. The endpoints are marked in orange, and the inner control points are in blue.}
\vspace{-0.5cm}
\label{B_bezier}
\end{figure}

In alignment with the setting outlined in \citet{wordAI}, we use the FreeType \cite{freetype} font library to extract the outlines of the specified letter. 
Subsequently, these outlines are converted into a closed curve composed of several cubic B\'ezier curves, as illustrated in Fig. \ref{B_bezier}, to achieve a coherent representation across different fonts and letters.
We iteratively subdivide the letter's B\'ezier segments until reaching a pre-defined threshold to ensure sufficient control points to represent semantic deformation.

\subsection{Score Distillation Sampling}
The objective of Score Distillation Sampling (SDS), originally introduced in the DreamFusion \cite{dreamfusion}, is to leverage pre-trained diffusion models' prior knowledge for the text-conditioned generation of different modalities \cite{nfsds}.
SDS optimizes the parameters $\theta$ of the parametric generator $\mathcal{G}$ (e.g., NeRF \cite{nerf}), ensuring the output of $\mathcal{G}$ aligns well with the prompt. 
For illustration, assuming $\mathcal{G}$ is a parametric image generator. 
First, an image $x=\mathcal{G}(\theta)$ is generated.
Next, a noise image $z_{\tau}(x)$ is obtained by adding a Gaussian noise $\epsilon$ at the diffusion process's $\tau$-th timestep: 
\begin{equation}\label{noise}
z_{\tau}(x)= \alpha_{\tau} x + \sigma_{\tau} \epsilon ,
\end{equation}
where $\alpha_{\tau}$, and $\sigma_{\tau}$ are diffusion model's noising schedule, and $\epsilon$ is a noise sample from the distribution $\mathcal{N}(0, 1)$.

For a pre-trained diffusion model $\epsilon_\phi$, the gradient of the SDS loss $\mathcal{L}_{SDS}$ is formulated as:
\begin{equation}\label{eq:sds_loss}
    \nabla_\phi \mathcal{L}_{SDS} = \left[ w(\tau)(\epsilon_\phi(z_{\tau}(x); y, \tau) - \epsilon) \frac{\partial x}{\partial \theta} \right] ,
\end{equation}
where $y$ is the conditioning input to the diffusion model and $w(\tau)$ is a weighting function.
The diffusion model predicts the noise added to the image $x$ with $\epsilon_\phi(z_{\tau}(x); y, \tau)$.
The discrepancy between this prediction and the actual noise $\epsilon$ measures the difference between the input image and one that aligns with the text prompt.
In this work, we adopt this strategy to extract the motion prior from the pre-trained text-to-video diffusion model \cite{modelscope}.

Since SDS is used within the pixel-based scenario, we utilize DiffVG \cite{diffvg} as a differentiable rasterizer. 
This allows us to convert our vector-defined content into pixel space in a differentiable way for applying the SDS loss.
\section{Method}
\label{sec:method}
Dynamic Typography focuses on animating individual letters within words based on the user's prompt. 
The letter is deformed to embody the word's semantic meaning and then brought to life by infusing motion based on the user's prompt.
To achieve visually appealing animations, we identify three crucial requirements for Dynamic Typography:

\begin{itemize}[leftmargin=*]
    \item \textbf{Legibility Preservation.} The deformed letter should remain legible in each frame during animation.
    \item \textbf{Semantic Alignment.} The letter should be deformed and animated in a way that aligns with the semantic information in the text prompt.
    \item \textbf{Temporal Consistency.} The deformed letter should move coherently while preserving a relatively consistent appearance in each animation frame.
\end{itemize}

\noindent\textbf{Problem Formulation:} 
The original input letter is initialized as a cubic B\'ezier curves control points set (Fig. \ref{B_bezier}), denoted as $P_{letter}=\{p_i\}_{i=1}^N=\{(x_i,y_i)\}_{i=1}^N \in \mathbb{R}^{N\times2}$, where $x,y$ refers to control points' coordinates in SVG canvas, and $N$ is the total number of control points of the indicated letter. 
The output video consists of k frames, each represented by a set of control points, denoted as $V=\{P^t\}_{t=1}^k \in \mathbb{R}^{k\times N \times 2}$, where $P^t$ is the control points for $t$-th frame.

Our goal is to learn the per-frame displacement to be added on the set of control point coordinates of the original letter's outline. 
This displacement represents the motion of the control points over time, creating the animation that depicts the user's prompt.
We denote the displacement for $t$-th frame as $\Delta P^t=\{\Delta p_i^t\}_{i=1}^{N} =\{(\Delta x_i^t,\Delta y_i^t)\}_{i=1}^{N} \in \mathbb{R}^{N \times2} $, where $\Delta p_i^t$ refers to the displacement of the $i$-th control point in the $t$-th frame.
The final video can be derived as $V=\{ P_{letter}+\Delta P^t\}_{t=1}^k$.

\begin{figure}[th!]
\centering
\vspace{-0.2cm}
\includegraphics[width=0.6\columnwidth]{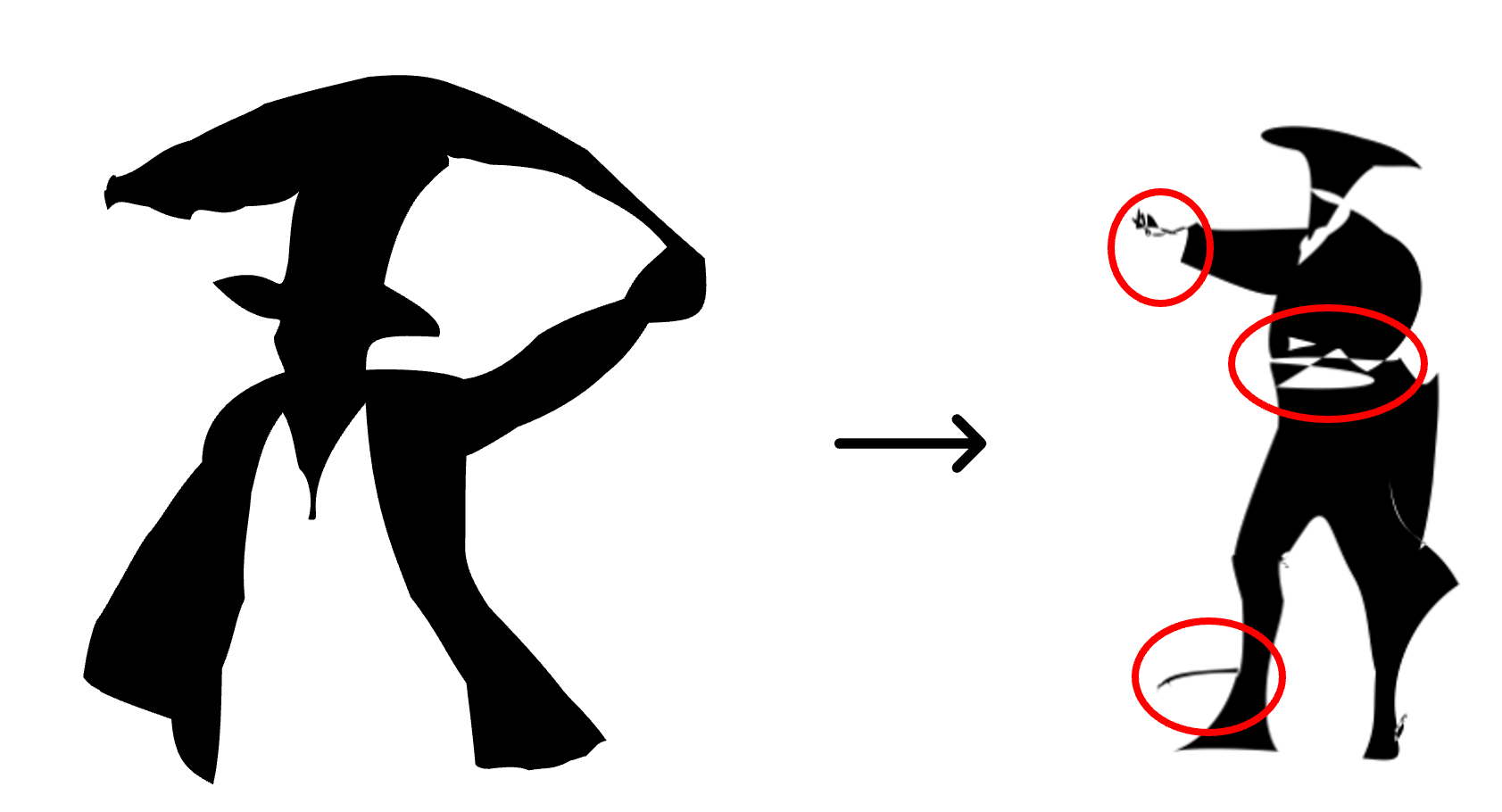}
\vspace{-0.25cm}
\caption{Illustration of the prior knowledge conflict issue. The left is the deformed ``R'' for ``BULLFIGHTER'' with prompt ``A bullfighter holds the corners of a red cape in both hands and waves it'' generated by \citet{wordAI}, the right is generated by \citet{livesketch} to animate the deformed letter with the same prompt. The mismatch in prior knowledge between separate models leads to significant appearance changes and severe artifacts, as highlighted by the red circles.} 
\vspace{-0.4cm}
\label{R_to_bullfighter}
\end{figure}

One straightforward strategy can be first deforming the static letter with existing method \cite{wordAI}, then utilizing an animation model \cite{livesketch} designed for vector graphics composed of B\'ezier curves to animate the deformed letter.
However, this non-end-to-end formulation suffers from conflicting prior knowledge. 
The deformed letter generated by the first model may not align with the prior knowledge of the animation model. 
This mismatch can lead the animation model to alter the appearance of the deformed letter, leading to considerable visual artifacts including distortion and inconsistency, see Fig. \ref{R_to_bullfighter}. 

Therefore, to ensure the coherence of the entire process, we propose an end-to-end framework that directly maps the original letter to the final animation, as illustrated in Fig. \ref{model_overview}.
To address the complexity of learning per-frame displacement that converts the letter into animation, we represent the video as a learnable base shape and per-frame motion added on the base shape (\S\ref{sec:base}). 
Additionally, we incorporate legibility regularization loss based on perceptual similarity to maintain letter legibility (\S\ref{sec:readability}). 
Finally, we introduce a mesh-based structure preservation regularization loss to ensure appearance and structure integrity between frames, mitigating inconsistent artifacts (\S\ref{sec:transition}).

\subsection{Base Field and Motion Field}\label{sec:base}
Learning the per-frame displacement that directly converts the input letter into animation frames is challenging.
Directly optimizing the displacement may lead to severe artifacts that degrade the quality of the animation, including distortion, flickering, and abrupt appearance changes in the adjacent frame.
Inspired by the dynamic NeRFs \cite{D-NeRF,hypernerf} and CoDeF \cite{codef}, we propose modeling the generated video in two neural displacement fields: the base field and the motion field, to address the complexity of this deformation.
Both fields are represented by coordinate-based Multilayer Perceptron (MLP). 
To better capture high-frequency variation and represent geometry information, we project the coordinates into a higher-dimensional space using the NeRF \cite{nerf} positional encoding:

{\footnotesize
\begin{equation}
\label{nerf_encoding}
\gamma(p) = \left( \sin(2^{0}\pi p), \cos(2^{0}\pi p), \ldots, \sin(2^{L-1}\pi p), \cos(2^{L-1}\pi p) \right)
\text{, }
\end{equation}
}

\noindent where $p$ refers to control point coordinates.
Further details can be found in the appendix.

The objective of the base field, denoted as $B$, is to learn a shared shape for every animation frame, serving as a base to infuse motion.
It is defined by a function $B: \gamma(P_{letter}) \rightarrow P_B$, which maps the original letter's control points coordinates $P_{letter}$ into base shapes' coordinates $P_B$, both in $\mathbb{R}^{N\times2}$. 

The motion field, denoted as $M$, encodes the correspondence between the control points in the base shape and those in each video frame. 
Inspired by CoDeF \cite{codef}, we represent the video as a 3D volume space, where a control point at $t$-th frame with coordinate $(x,y)$ is represented by $(x,y,t)$.
We duplicate the shared base shape $k$ times across $k$ frames and convert the coordinates into 3D volume space, written as $P_B': \{(P_B,t)\}_{t=1}^k$.
The motion field is defined as a function $M: \gamma(P_B')\rightarrow P_V$ that maps control points from the base shape to their corresponding locations $P_V \in \mathbb{R}^{k\times N \times 2}$ in each video frame.

To better model motion, we represent $P_V$ as $P_B+\Delta P$, focusing on learning the per-frame displacements $\Delta P=\{\Delta P^t\}_{t=1}^k$ to be applied on the base shape.
Following \citet{livesketch}, we decompose the motion into global motion (modeled by an affine transformation matrix shared by all control points of an entire frame) and local motion (predicted for each control point separately).
Consider the $i$-th control point on the base shape with coordinate $(x_{B,i}, y_{B,i})$, its displacement on $t$-th frame $\Delta {p}_i^t$ is summed by its local and global displacement:

{\footnotesize
\begin{equation}
\Delta {p}_i^t = \Delta p_i^{t, local} + \Delta {p}_i^{t, global},
\end{equation}
\vspace{-1ex}
\begin{equation}
\setlength{\arraycolsep}{0.1pt}
\begin{bmatrix}
    \Delta p_i^{t, global}\\
    1
\end{bmatrix} = 
\left[
\begin{array}{ccc}
s_x & sh_x s_y & d_x \\
sh_y s_x & s_y & d_y \\
0 & 0 & 1
\end{array}
\right]
\left[
\begin{array}{ccc}
\cos\theta & \sin\theta & 0 \\
-\sin\theta & \cos\theta & 0 \\
0 & 0 & 1
\end{array}
\right]
\begin{bmatrix}
x_{B,i} \\
y_{B,i} \\
1
\end{bmatrix}
-
\begin{bmatrix}
x_{B,i} \\
y_{B,i} \\
1
\end{bmatrix}
, 
\end{equation}
}

\noindent where $\Delta p_i^{t, local}$ and all elements in the per-frame global transformation matrix are predicted by the MLP in the motion field.

\begin{figure}[th!]
\centering
\includegraphics[width=0.25\textwidth]{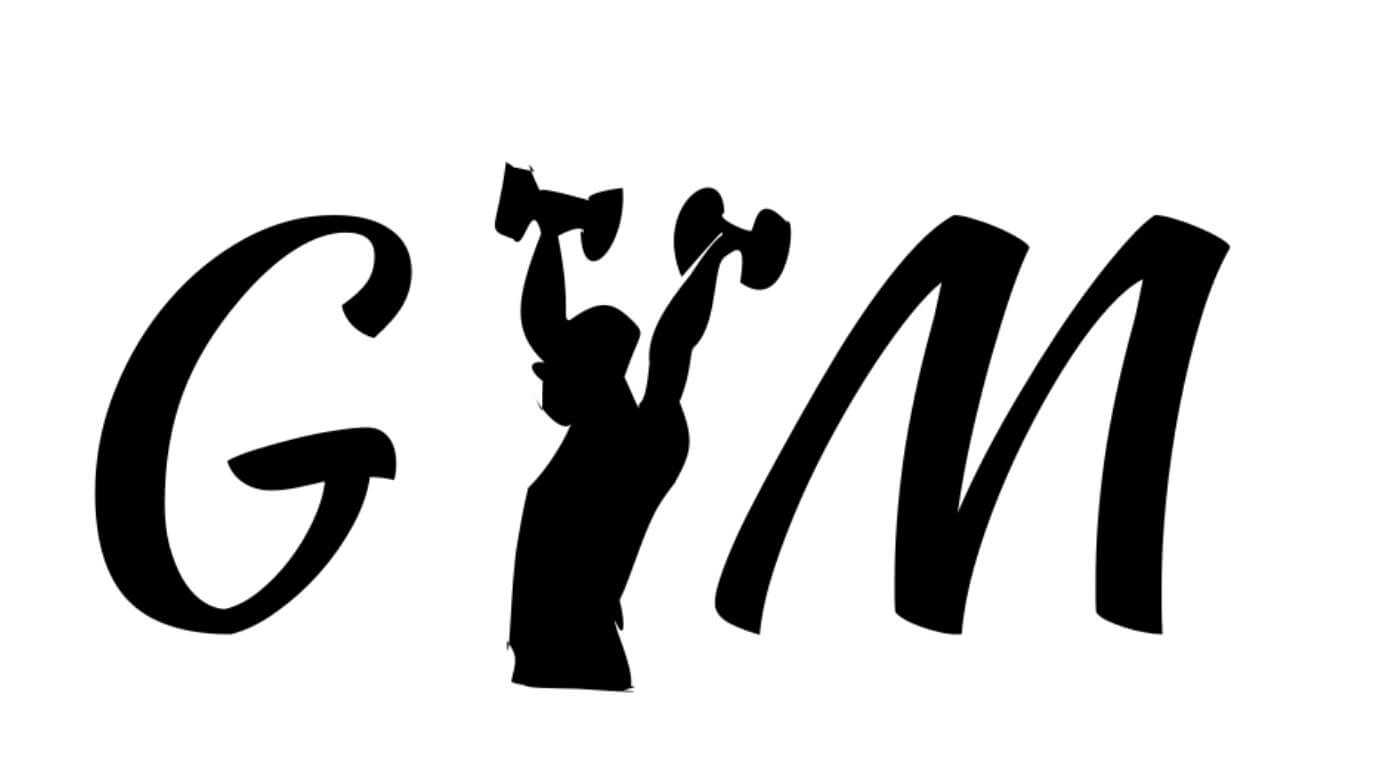}
\caption{Base shape of ``Y'' for ``GYM'' with prompt ``A man doing exercise by lifting two dumbbells in both hands.''}
\label{base}
\end{figure}

To train the base field and motion field, we distill prior knowledge from the large-scale pretrained text-to-video model using SDS computed in Eq. \ref{eq:sds_loss}. 
At each training iteration, we use a differentiable rasterizer \cite{diffvg}, denoted as $R$, to render our predicted control points set $P_V$ into a rasterized video (pixel format video). 
We select a diffusion timestep $\tau$, draw a sample from a normal distribution for noise $ \epsilon \sim \mathcal{N}(0, 1) $,  and then add the noise to the rasterized video. 
The video foundation model denoise this video, based on the user prompt describing a motion pattern closely related to the word's semantic meaning (e.g. ``A camel walks steadily across the desert'' for ``M'' in ``CAMEL''). 
We jointly optimize the base field and motion field using the SDS loss to generate videos aligned with the desired text prompt.
The visualized base shape demonstrates alignment with the prompt's semantics, as shown in Fig. \ref{base}.

\subsection{Legibility Regularization}\label{sec:readability}

A critical requirement for Dynamic Typography is ensuring the animations maintain legibility. 
For example, for ``M'' in ``CAMEL'', we hope the ``M'' takes on the appearance of a camel while being recognizable as the letter ``M''. 
When employing SDS loss for training, the text-to-video foundation model's prior knowledge naturally deforms the letter's shape to match the semantic content of the text prompt. 
However, this significant appearance change compromises the letter's legibility throughout the animation.

Thus, we propose a regularization term that enforces the letter to be legible, working alongside the SDS loss to guide the optimization process. 
Specifically, we leverage Learned Perceptual Image Patch Similarity (LPIPS) \cite{lpips} as a loss to regularize the perceptual distance between the rasterized images of the base shape $R(P_B)$ and the original letter $R(P_{letter})$: 
\begin{equation}
   \mathcal{L}_{\text{legibility}} = \text{LPIPS}\left(R\left(P_B\right),R\left(P_{letter}\right)\right).
\end{equation} 
Benefiting from our design, we only need to apply this LPIPS-based legibility regularization term to the base shape, and the motion field will automatically propagate this legibility constraint to each frame. 

\subsection{Structure Preservation Regularization}\label{sec:transition}
\begin{figure}[htbp]
  \centering
  \begin{subfigure}[b]{0.24\columnwidth}
    \includegraphics[width=\linewidth]{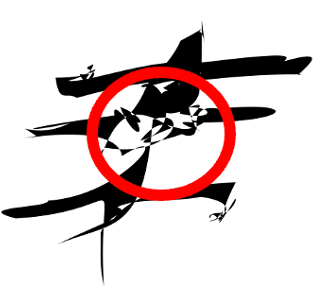} 
    \caption{frame 1}
  \end{subfigure}
  \begin{subfigure}[b]{0.24\columnwidth}
    \includegraphics[width=\linewidth]{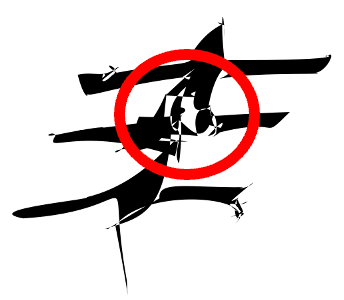} 
    \caption{frame 2}
  \end{subfigure}
  \begin{subfigure}[b]{0.24\columnwidth}
    \includegraphics[width=\linewidth]{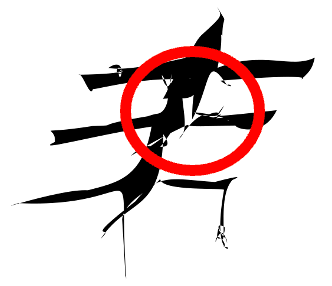} 
    \caption{frame 3}
  \end{subfigure}
    \begin{subfigure}[b]{0.24\columnwidth}
    \includegraphics[width=\linewidth]{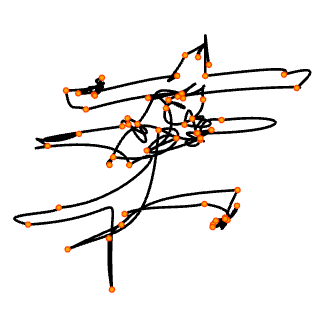} 
    \caption{frame 1 vis.}
    \label{vis-jet}
  \end{subfigure}
\caption{Adjacent frames of animation for letter ``E'' in ``JET''. A large area of alternating black and white ``holes'' occur within each frame, as highlighted within the red circles, causing severe flickering between the adjacent frames. (d) is the visualization of frame 1, highlighting the control points and the associated B\'ezier curves. The illustration reveals frequent intersections among the B\'ezier curves leading to the flickering artifacts.}
  \label{adjacent-jet123}
\end{figure}

The optimization process alters the positions of control points, sometimes leading to complex intersections between B\'ezier curves, as illustrated in Fig. \ref{vis-jet}.
The rendering of Scalable Vector Graphics (SVG) adheres to the non-zero rule or even-odd rule \cite{even-odd}, which determines the fill status by drawing an imaginary line from the point to infinity and counting the number of times the line intersects the shape's boundary. 
The frequent intersections between B\'ezier curves complicate the boundary, leading to alternating black and white ``holes'' within the image. 
Furthermore, these intersections between B\'ezier curves vary abruptly between adjacent frames, leading to severe flickering effects that degrade animation quality, see Fig. \ref{adjacent-jet123}.
In addition, the unconstrained degrees of freedom in motion could alter the appearance of the base shape, leading to noticeable discrepancies in appearance between adjacent frames and temporal inconsistency.

\begin{figure}[th!]
\centering
\includegraphics[width=0.96\columnwidth]{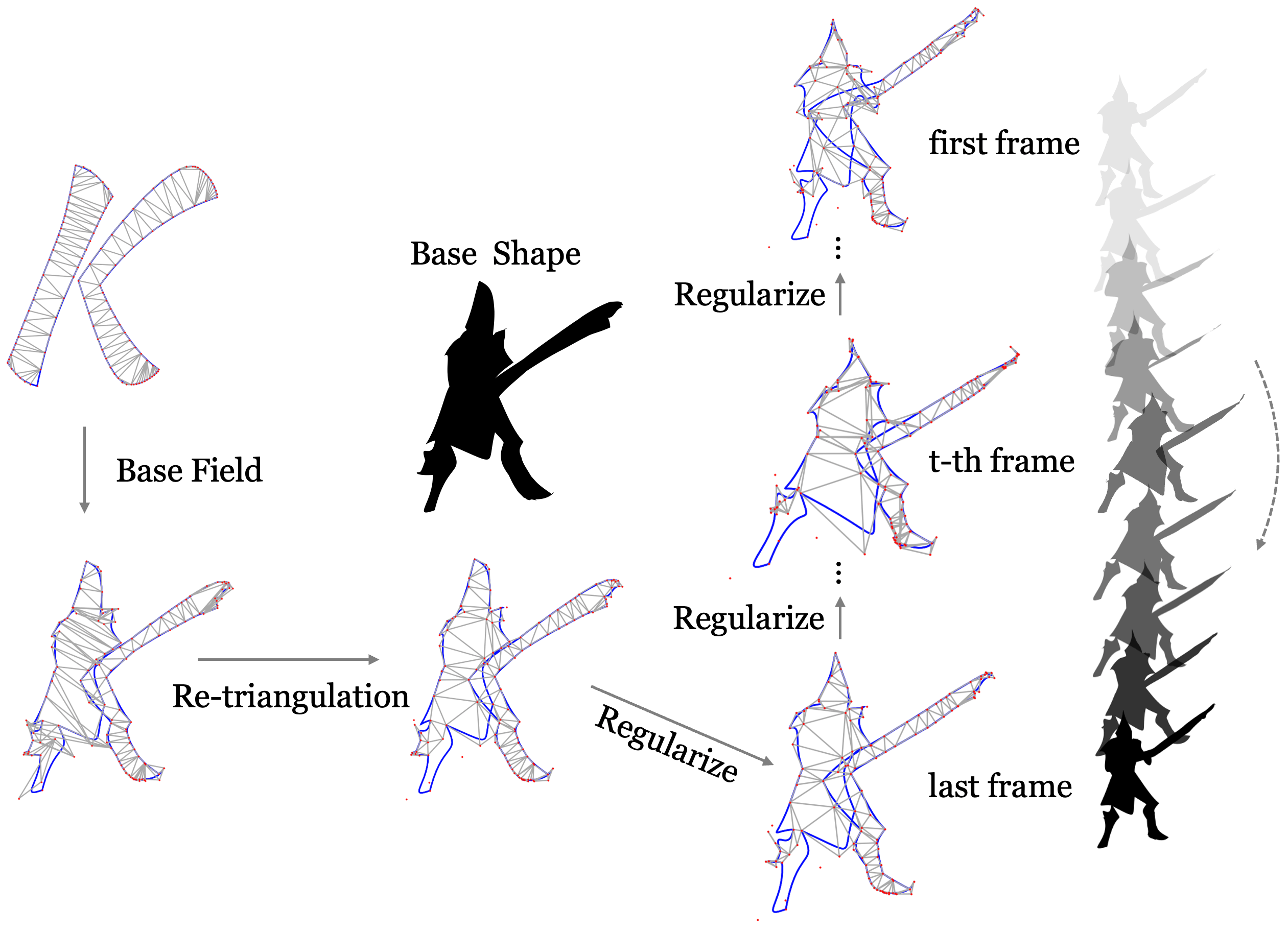}
\caption{Illustration of the Mesh-based structure preservation. We first apply this regularization between the base shape and the input letter. We propagate the structural constraint to every frame by regularizing the last frame with the base shape and regularizing every frame with its next frame.}
\label{triangulation_framework}
\vspace{-0.2cm}
\end{figure}

To address these issues, we adopt Delaunay Triangulation \cite{delaunay,delaunay2} on the base shape based on control points, as shown in Fig. \ref{triangulation_framework}. 
By maintaining the structure of the triangular meshes, we prevent frequent intersections between B\'ezier curves, while preserving the relative consistency of local geometry information across adjacent frames.

The whole regularization process is illustrated in Fig. \ref{triangulation_framework}.
We employ the angle variation \cite{wordAI} of the corresponding triangular meshes in adjacent frames as regularization:
\begin{equation}
    \frac{1}{ k\times m} \sum_{t=1}^{k} \sum_{i=1}^{m} \|T_{i,t+1} - T_{i,t} \|_2^2,
\end{equation}
where $m$ refers to the total number of triangular meshes in each frame, $T_{i,t}\in \mathbb{R}^3$ refers to the corresponding angles in $i$-th triangular mesh in the $t$-th frame.  
Particularly, the $(k+1)$-th frame refers to the base shape. 
Consequently, the structural constraint with the base shape is propagated to every frame, allowing the preservation of the geometric structure throughout the animation. 
Furthermore, to ensure the base shape is geometrically similar to the input letter, we apply the same triangulation-based constraints between the base shape and the letter. $\mathcal{L}_{\text{structure}}$ can be formulated as:
\begin{equation}
\begin{split}
\mathcal{L}_{\text{structure}} = & \lambda_1 \cdot \frac{1}{m} \sum_{i=1}^{m} \| T_{i,\text{letter}} - T_{i,B} \|_2^2 \\
& + \lambda_2 \cdot \frac{1}{k \times m} \sum_{t=1}^{k} \sum_{i=1}^{m} \| T_{i,t+1} - T_{i,t} \|_2^2,
\end{split}
\end{equation}
where $\lambda_1$ and $\lambda_2$ are weight hyperparameters .

We find that this angle-based na\"ive approach effectively maintains the triangular structure, thereby alleviating the frequent intersections of the B\'ezier curves and preserving a relatively stable appearance across different frames without significantly affecting the motion liveliness.

\begin{figure*}[h!t]
\centering
\includegraphics[width=1.0\textwidth]{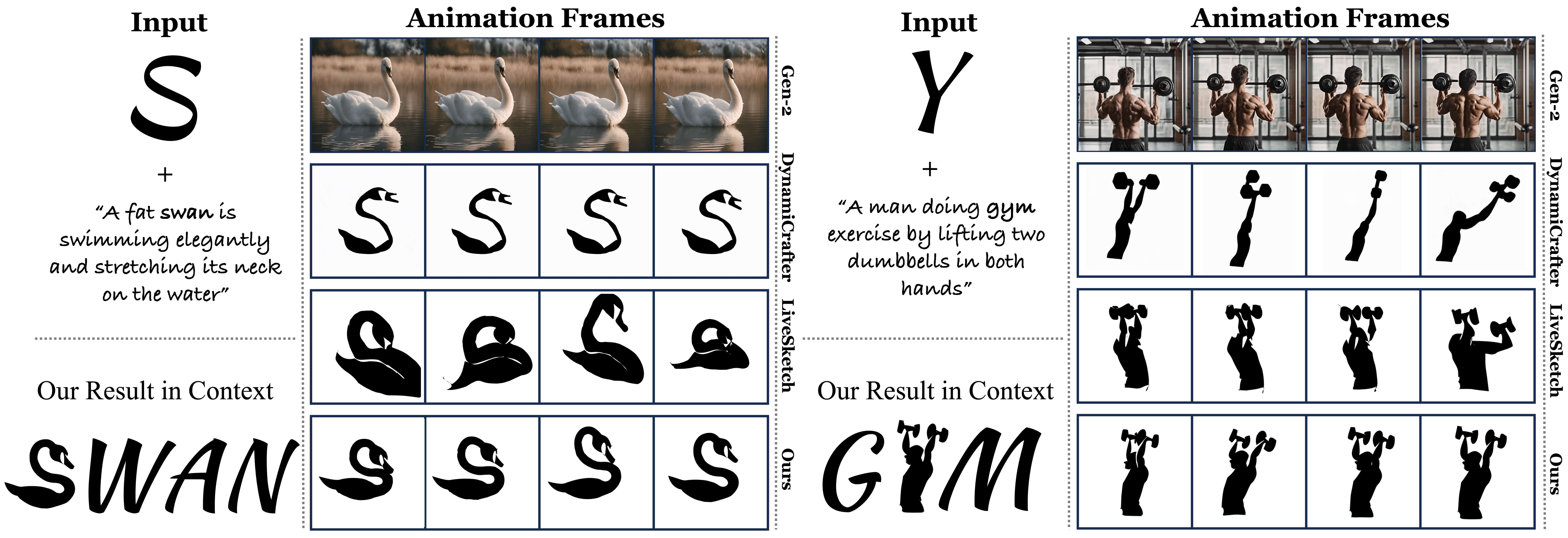}
\caption{Visual comparisons between the baselines and our model. Text-to-video model (Gen-2) generates colorful images but fails to maintain the shape of the original letter. The pixel-based image-to-video model (DynamiCrafter) produces results with minimal, sometimes unreasonable motion. The general vector animation model (LiveSketch) struggles to preserve legibility or maintain a consistent appearance.}
\label{fig:comparison}
\vspace{-0.3cm}
\end{figure*} 

\section{Experiments}
We create a dataset via a workshop to evaluate our method's ability. 
The dataset covers Dynamic Typography samples for all letters in the alphabet, featuring a variety of elements such as animals, humans, and objects, in a total of 33 samples.
Each sample includes a word, a specific letter within the word to be animated, and a concise text prompt describing the desired animation. 
For each sample, a video with 24 frames will be generated.
Each sample takes 1000 epochs for optimization, about 40 minutes on a H800 GPU. 

To illustrate our method's capabilities, we present some generated results in Fig. \ref{fig:teaser}. 
These animations vividly bring the specified letter to life while adhering to the prompt and maintaining the word's readability.
The implementation details and more results can be found in our appendix and supplementary materials.

\subsection{Comparisons}
We compare our method with approaches from two distinct categories: the pixel-based strategies leveraging either text-to-video or image-to-video methods, and the vector-based animation method.

Within the pixel-based scenario, we compare our model against the leading text-to-video generation models Gen-2 \cite{gen-2} (ranked first in the EvalCrafter \cite{evalcrafter} and VBench \cite{vbench} benchmark) – a commercial web-based tool, and DynamiCrafter \cite{dynamicrafter}, the state-of-the-art model for image-to-video generation conditioned on text. 
For text-to-video generation, we append the prompt with ``which looks like a letter $\beta$,'' where $\beta$ represents the specific letter to be animated. 
In the image-to-video case, we use the stylized letter generated by the Word-as-Image \cite{wordAI} as the conditioning image. 
Within the vector-based scenario, we utilize LiveSketch \cite{livesketch} as a framework to animate vector images. To ensure a fair comparison, we condition the animation on the stylized letter generated by the Word-as-Image \cite{wordAI} as well.
For fairness, all animations to be compared are composed of 24 frames, each lasting 3 seconds, and are rendered at a resolution of $256 \times 256$.

\textbf{Qualitative Comparison}
We present two samples for visual comparison in Fig. \ref{fig:comparison}.
For more comparisons and full videos, please check our supplementary materials.
While achieving high resolution and realism, Gen-2 struggles to generate frames that keep the letter's shape, which greatly harms the legibility.
With DynamiCrafter, the ``SWAN'' animation exhibits minimal movement, while the ``GYM'' animation features unrealistic motion that deviates from the user's prompt.
Although LiveSketch can depict the user's prompt through animation, it sacrifices legibility. Also, the animated letter fails to maintain a stable appearance throughout the animation, as demonstrated in the ``SWAN'' example.
Our model strikes a balance between prompt-video alignment and letter legibility.
It consistently generates animations that adhere to the user's prompt while preserving the original letter's form. 
This allows the animation to seamlessly integrate within the original word, as showcased by the in-context results in Fig. \ref{fig:comparison}.

\begin{table}[!h]
    \centering 
    \SetTblrInner{rowsep=0.0pt}      
    \SetTblrInner{colsep=3.0pt}      
    \begin{tblr}{
        cells={halign=c,valign=m},   
        column{1}={halign=l},        
        hline{1,3,7}={1-7}{},        
        hline{1,3,7}={1.0pt},          
        vline{2,3}={1-7}{},          
        cell{1}{1}={r=2}{},      
    }
    \ Method & Perceptual & Text-to-Video  \\
     & Input Conformity $\left(\uparrow\right)$ & Alignment$\left(\uparrow\right)$ \\
    Gen-2 & $0.1495$ & $\textcolor{red}{\mathbf{23.3687}}$ \\
    DynamiCrafter & $0.5151$ & $17.8124$ \\
    LiveSketch & $0.4841$ & $20.2402$ \\
    Ours & $\textcolor{red}{\mathbf{0.5301}}$ & $21.4391$ \\
    \end{tblr}
    \caption{Quantitative results between the baselines and our model. The best score for each metric is highlighted in red. Our model gets the best score in PIC, and second-best score in text-to-video alignment, indicating a balance between faithfully representing the user's prompt and maintaining the legibility.}\label{tab:comparison}
\end{table}

\textbf{Quantitative Comparison} 
Tab. \ref{tab:comparison} presents the quantitative evaluation results. 
We employ two metrics, Perceptual Input Conformity (PIC) and Text-to-Video Alignment.
Following DynamiCrafter \cite{dynamicrafter}, we compute Perceptual Input Conformity (PIC) using DreamSim's \cite{dreamfusion} perceptual similarity metric between each output frame and the input letter, averaged across all frames. 
This metric assesses how well the animation preserves the original letter's appearance.
To evaluate the alignment between the generated videos and their corresponding prompts (Text-to-Video Alignment), we leverage the X-CLIP score \cite{xclip}, which extends CLIP \cite{clip} to video recognition, to obtain frame-wise image embeddings and text embeddings. 
The average cosine similarity between these embeddings reflects how well the generated videos align with the corresponding prompts. 

While Gen-2 achieves the highest text-to-video alignment score, it severely suffers in legibility preservation (lowest PIC score). 
Conversely, our model excels in PIC, indicating the effectiveness in maintaining the original letter's form. 
Our model also achieves the second-best text-to-video alignment score, indicating it faithfully depicts the animation prompt while preserving legibility.


\subsection{Ablation Study}
\begin{figure}[!h]
\centering
\includegraphics[width=1\columnwidth]{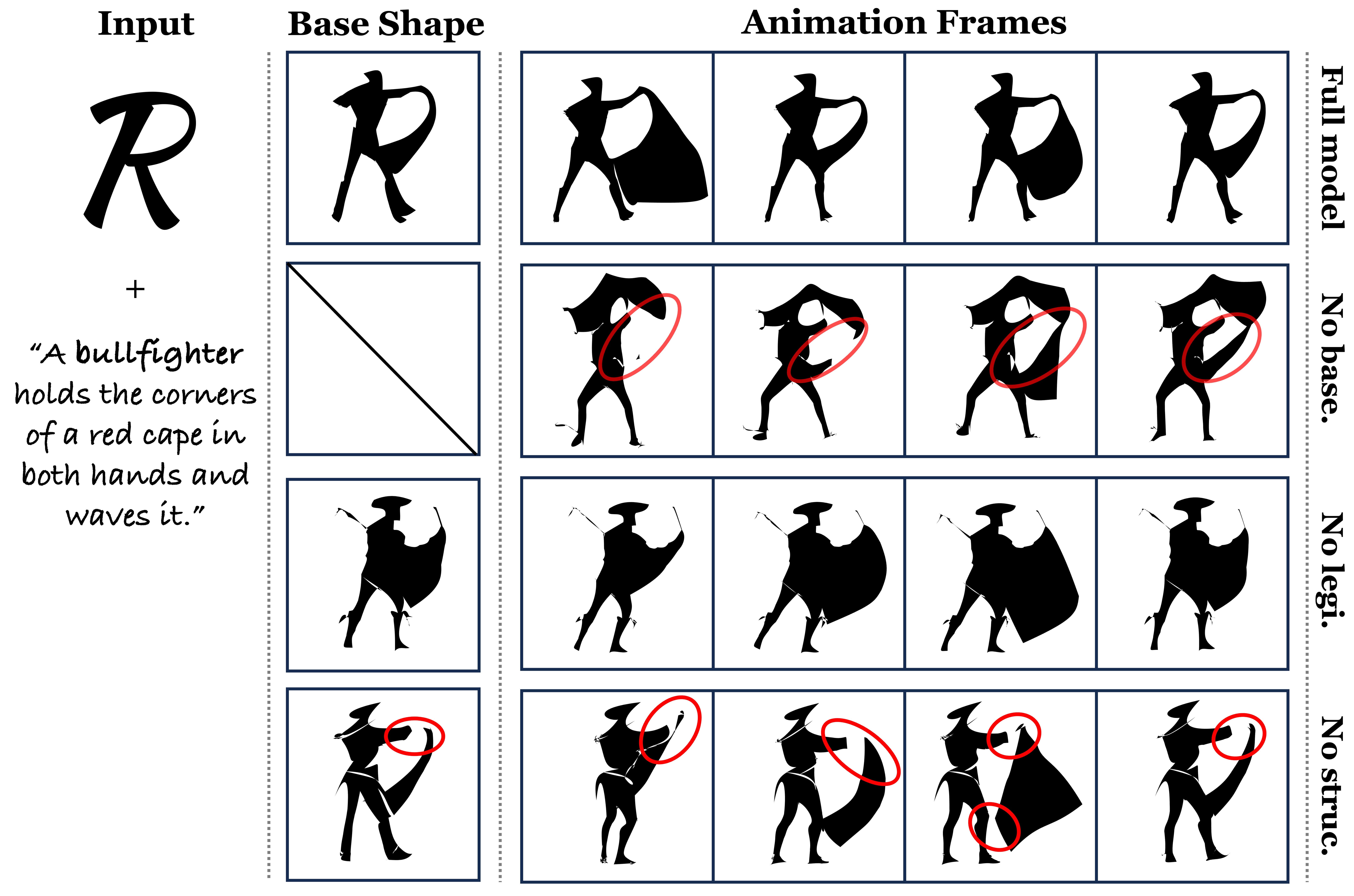}
\vspace{-0.2cm}
\caption{Visual comparisons of ablation study. Some artifacts are highlighted in red cycles. Removing base shape or structure preservation regularization results in shape deviation and inconsistent artifacts. Without legibility regularization, each animation frame loses the letter ``R'' shape.}
\label{fig:ablation}
\end{figure}

We conduct ablations to analyze the contribution of each component in our proposed method: learnable base shape, legibility regularization, and mesh-based structure preservation regularization.
Visual results in Fig. \ref{fig:ablation} showcase the qualitative impact of removing each component. 
Quantitative results in Tab. \ref{tab:ablation} further confirm their effectiveness.

In addition to Perceptual Input Conformity (PIC) and X-CLIP score (Text-to-Video Alignment), we employ warping error to assess temporal consistency, following EvalCrafter \cite{evalcrafter}. 
This metric estimates the optical flow between consecutive frames using the RAFT model \cite{raft} and calculates the pixel-wise difference between the warped image and the target image. 
The lower warping error indicates smoother and more temporally consistent animations.

The calculation of legibility and structure preservation regularization loss involves the base shape. 
Hence, when removing the learnable base shape, the legibility loss $\mathcal{L}_{\text{legibility}}$ is computed between every output frame and the input letter, and the structure preservation loss $\mathcal{L}_{\text{structure}}$ is only applied between every pair of consecutive frames.

\begin{table}[!h]
    \centering \small
    \SetTblrInner{rowsep=0.0pt}      
    \SetTblrInner{colsep=3.0pt}      
    \begin{tblr}{
        cells={halign=c,valign=m},   
        column{1}={halign=l},        
        hline{1,2,6}={1-7}{},        
        hline{1,2,6}={1.0pt},          
        vline{2,3,4}={1-7}{},          
    }
    \   Method 
   & Warping Error$\left(\downarrow\right)$ & PIC $\left(\uparrow\right)$  & T2V Align. $\left(\uparrow\right)$   \\
    Full Model  & $0.01645$ & $\textcolor{red}{\mathbf{0.5310}}$ & $\textcolor{red}{\mathbf{21.4447}}$ \\
    No Base Shape & $0.03616$ & $0.5178$ & $20.0568$ \\
    No Legibility & $\textcolor{red}{\mathbf{0.01561}}$ & $0.4924$ & $20.2857$ \\
    No Struc. Pre. & $0.01777$ & $0.4906$ & $20.6285$ \\
    \end{tblr}
    \captionof{table}{%
    Quantitative results of the ablation study. The best score for each metric is highlighted in red. The full model gets the best in PIC, text-to-video alignment, and second-best in optical flow warping error, indicating the effectiveness of each module.} \label{tab:ablation}
    \vspace{-0.2cm}
\end{table}

\noindent\textbf{Base Shape:}
As observed in Fig. \ref{fig:ablation} (row 2), removing the shared learnable base shape results in inconsistent animations. 
Specifically, as highlighted by the red circle, the appearance of the bullfighter deviates significantly between frames, harming legibility. 
The finding is also supported by Tab. \ref{tab:ablation} (row 2), where removing the base shape results in significant degradation under all three metrics.

\noindent\textbf{Legibility Regularization:}
Without the perceptual regularization on the base shape, the base shape struggles to preserve legibility. 
As a result, each animation frame loses the letter ``R'' shape in Fig. \ref{fig:ablation} (row 3), leading to lower PIC in Tab. \ref{tab:ablation} (row 3).

\noindent\textbf{Structure Preservation Regularization:}
Removing mesh-based structure preservation allows the base shape's structure to deviate from the original letter, causing the discontinuity between the bullfighter and cape in the base shape and all frames, as highlighted in Fig. \ref{fig:ablation} (row 4). 
Without this regularization term, the animation shows inconsistent appearances across different frames, degrading legibility and leading to the lowest PIC in Tab. \ref{tab:ablation} (row 4).

\subsection{Generalizability}
\begin{figure}[!ht]
\centering
\includegraphics[width=1\columnwidth]{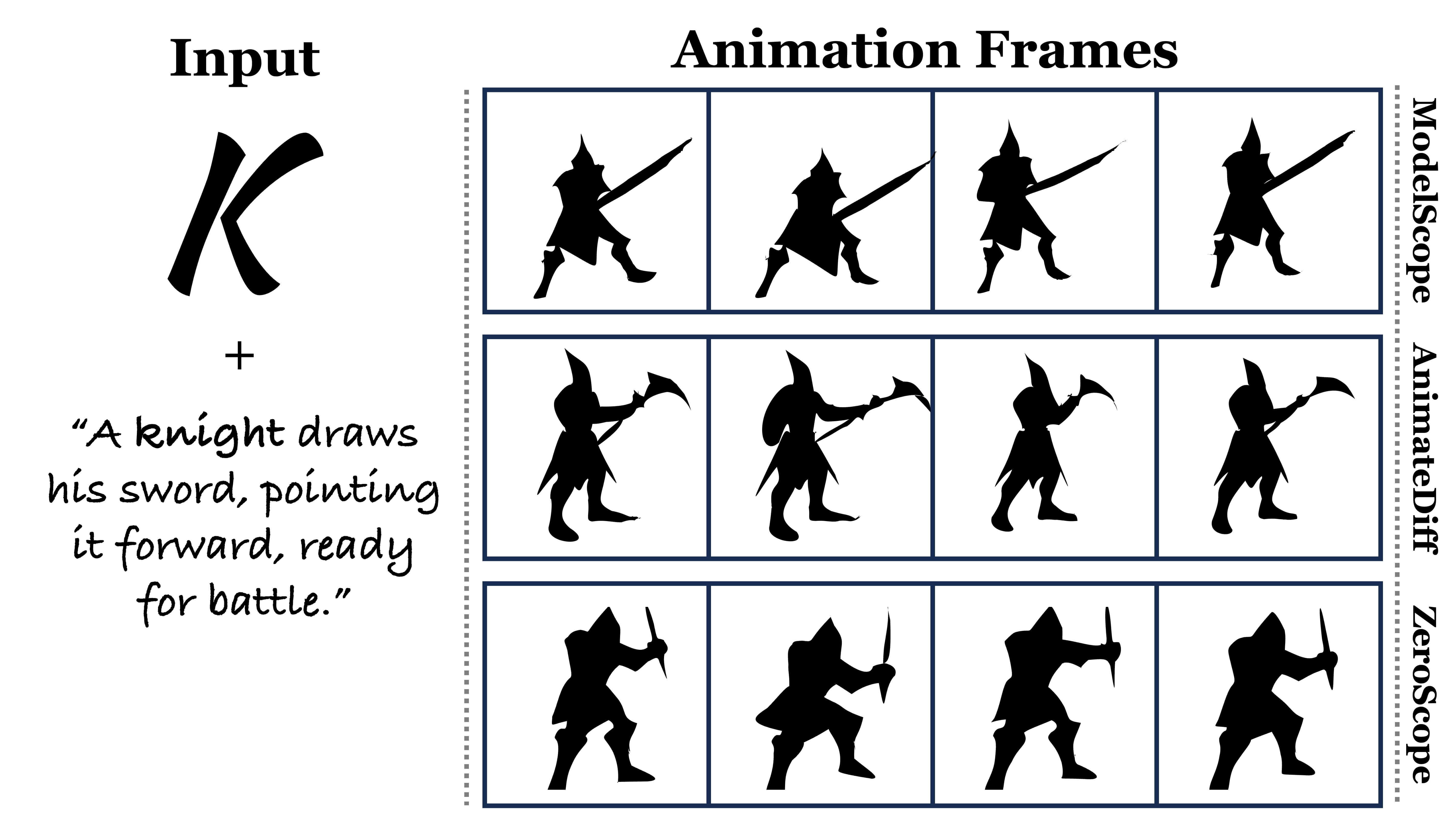}
\caption{Visual results of the same animation sample (``K'') using different text-to-video base models. All generated animations accurately depict the user's prompt and maintain the ``K'' shape, showing the generalizability of our methods across different text-to-video foundation models.}
\label{fig:general}
\end{figure}

Our framework, leveraging Score Distillation Sampling (SDS), achieves generalization across various diffusion-based text-to-video models.
To demonstrate this, we apply different base models for computing $\mathcal{L}_{\text{SDS}}$, including the 1.7-billion parameter text-to-video model from ModelScope \cite{modelscope}, AnimateDiff \cite{animatediff}, and ZeroScope \cite{VideoFusion}. 
Fig. \ref{fig:general} presents visual results for the same animation sample (``K'' in ``Knight'') with each base model. 

While the animation exhibits deformations and animation styles unique to each model, all animations accurately depict the user's prompt and maintain the ``K'' shape. 
This showcases the generalizability of our method. 
Hence, future advancements in text-to-video models with stronger prior knowledge will benefit our approach.


\section{Conclusion}
We propose a text animation scheme, termed ``Dynamic Typography'', that deforms letters to convey semantic meaning and animates them vividly based on user prompts.
To automate text animation generation, we propose an end-to-end optimization-based framework that leverages the video diffusion prior.
Our method faithfully depicts the user's prompt in animation while preserving the input letter's legibility, and is generalizable to different fonts, prompts, and languages.
Nevertheless, there remain several limitations. 
First, the motion quality can be bounded by the video foundation model, which may be unaware of specific motions in some cases. 
Luckily, our framework is model-agnostic, facilitating integration with future diffusion-based video foundation model advancements.
Besides, challenges arise when user-provided text prompts deviate significantly from original letter shapes, complicating the model's ability to strike a balance between generating semantic-aware vivid motion and preserving the legibility of the original letter.
We hope that our work can open the possibility for further research of semantic-aware text animation that incorporates the rapid development of video generation models.

\clearpage
{
\small
\bibliographystyle{ieeenat_fullname}
\bibliography{ref.bib}
}

\clearpage
\appendix
\section*{Appendix}

\section{Implementation Details}
The base field and motion field are jointly optimized. Following \citet{livesketch}, we interleavely optimize local motion and global motion. 
Adam optimizer is adopted, and the learning rates are set to be $5e-3$, $5e-3$, and $1e-3$ to learn the base field, local motion, and global motion.
For the global motion, we set the scaling factor to be $2.0$, $1e-2$, $5e-2$, and $1e-1$ for translation, rotation, scale, and shear.

We set the regularization weight to be $5e3$ for the $\mathcal{L}_{\text{legibility}}$ and $1e3$, $1e4$ for the $\lambda_1$, $\lambda_2$ in $\mathcal{L}_{\text{structure}}$.
We observe that $\mathcal{L}_{\text{legibility}}$ often plays a dominant role. When perceptual regularization is employed from the start, the base shape typically retains the original letter's form, preventing any semantic deformations.
Hence, following \citet{wordAI}, we gradually increase the weight of $\mathcal{L}_{\text{legibility}}$ to make its effects after semantic deformation has taken place. 

We use the text-to-video-ms-1.7b model in ModelScope \cite{modelscope, VideoFusion} for the diffusion backbone. 
We apply augmentations including random crop and random perspective to all video frames. 
We intend to make our code available to release all the details and support further research.

\section{Frequency-based Encoding and Annealing}
NeRF \cite{nerf} has highlighted that a heuristic application of sinusoidal functions to input coordinates, known as ``positional encoding'', enables the coordinate-based MLPs to capture higher frequency content, as denoted by: 

{\footnotesize
\begin{equation}
\label{nerf_encoding}
\gamma(p) = \left( \sin(2^{0}\pi p), \cos(2^{0}\pi p), \ldots, \sin(2^{L-1}\pi p), \cos(2^{L-1}\pi p) \right)
\text{, }
\end{equation}
}
where $p$ refers to the point coordinates.

We found that this property also applies to our MLPs that use coordinates of control points as input.
This allows the MLPs in the base and motion field to more effectively represent high-frequency information, corresponding to the detailed geometric features.
Additionally, when using coordinate-based MLPs to model motion, a significant challenge is capturing both minute and large motions. 
Following Nerfies \cite{nerfies}, we employ a coarse-to-fine strategy that initially targets low-frequency (large-scale) motion and progressively refines the high-frequency (localized) motions. 
Specifically, we use the following formula to apply weights to each frequency band $j$ in the positional encoding of the MLPs within the motion field.

\begin{equation}
 w_j(\alpha) = \frac{1 - \cos(\pi \cdot \text{clamp}(\alpha - j, 0, 1))}{2},
\end{equation}
where  $\alpha(t) = \frac{Lt}{N}$, $t$ is the current training iteration, and \(N\) is a hyper-parameter for when \(\alpha\) should reach the maximum number of frequencies \(L\).

\begin{figure}[H]
\centering
\includegraphics[width=1\columnwidth]{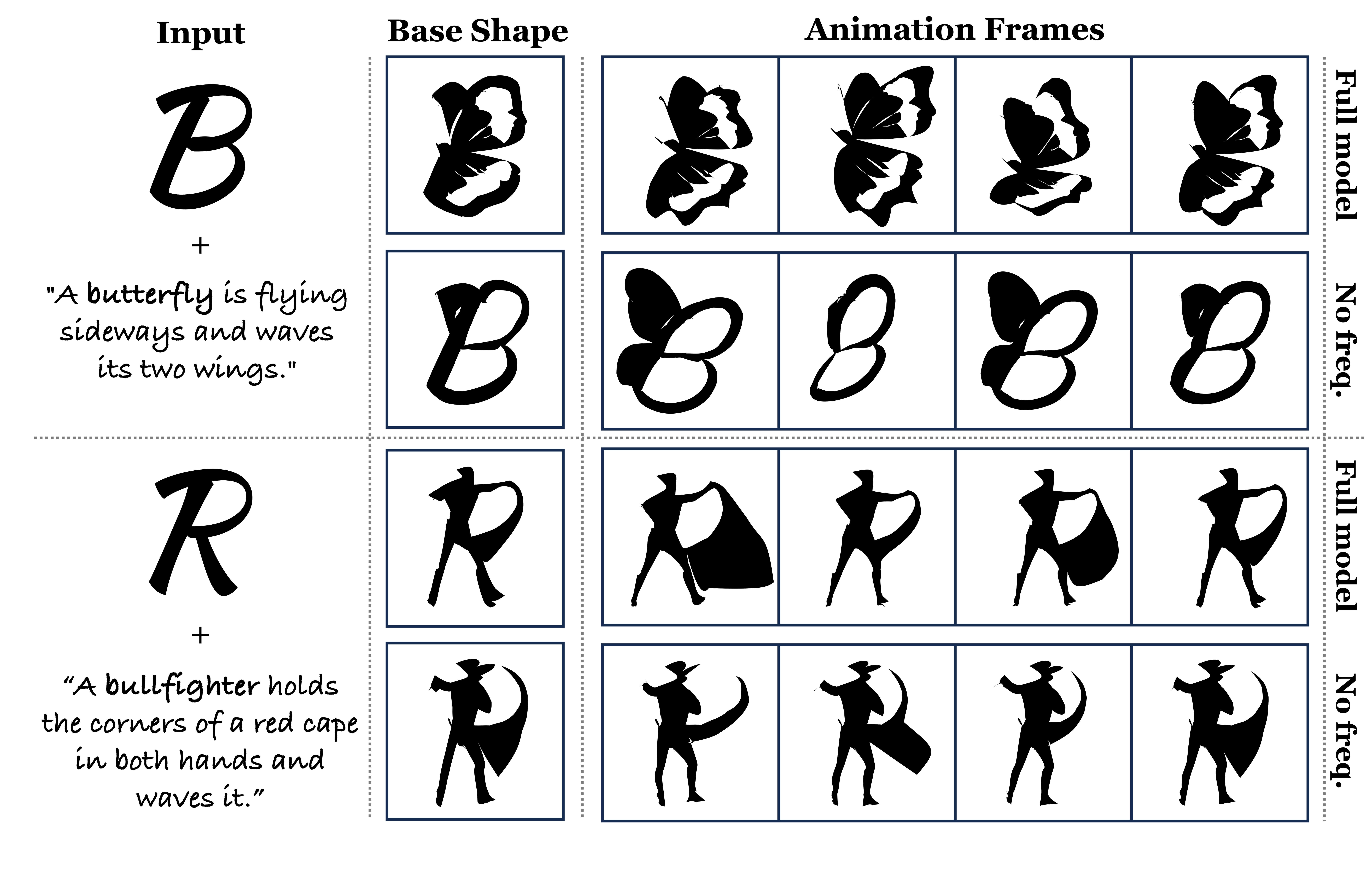}
\caption{Visual comparison with and without frequency encoding and annealing. The geometry and motion quality degrades when removing annealed frequency-based encoding.}
\label{fig:ablation_freq}
\end{figure} 

Fig. \ref{fig:ablation_freq} shows the visual ablation comparisons with and without the frequency encoding and annealing. 
When removing annealed frequency-based encoding, the geometry and motion quality of the generated animation suffer.
To be specific, the butterfly animation in Fig. \ref{fig:ablation_freq} (row 2) exhibits a lack of geometry details, and the bullfighter animation in Fig. \ref{fig:ablation_freq} (row 4) shows unreasonable motion, leading to the degradation of the animation quality.

\section{Effect Analysis of Control Points}
\begin{figure}[H]
\centering
\includegraphics[width=1\columnwidth]{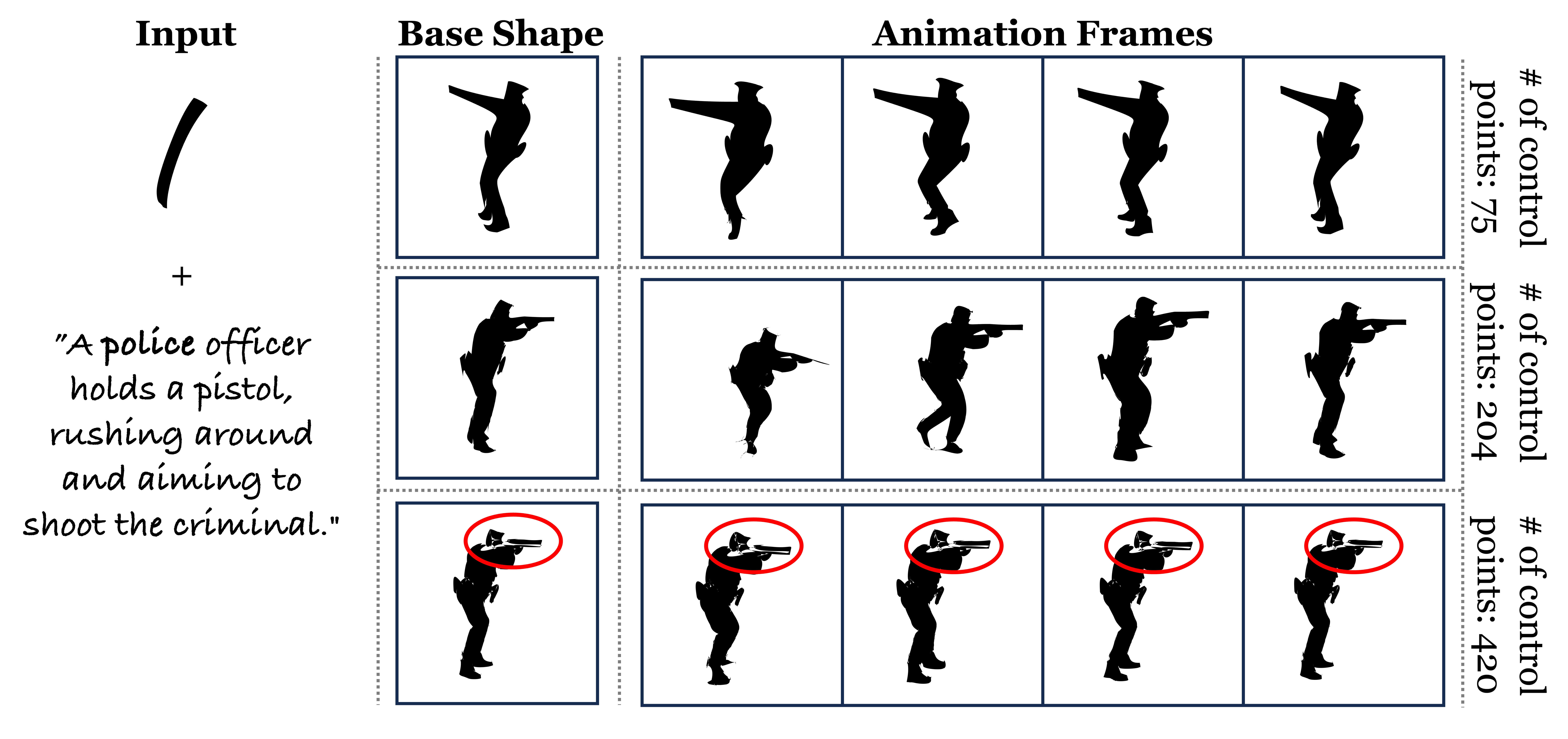}
\caption{The effect of the number of control points. The first row displays the result generated with the default number of control points extracted by FreeType \cite{freetype}, which is 75. In the second and third rows, the number of control points is increased to 204 and 420 respectively.}
\label{fig:cp}
\end{figure}

By adjusting the number of control points on the B\'ezier curves, we can alter the appearance of the generated text animations, thus producing more diverse outcomes. 
Generally, increasing the number of control points enriches the geometric details, as shown in Fig. \ref{fig:cp} (row 2). 
However, including too many control points may lead to self-intersection of the B\'ezier curves, causing abrupt and frequent changes of black and white regions, as highlighted by the red circles in Fig. \ref{fig:cp} (row 3). 
Users can choose the sample they desire based on their preferences.

\clearpage

\begin{figure*}[!htb]
\centering
\includegraphics[width=1.0\textwidth]{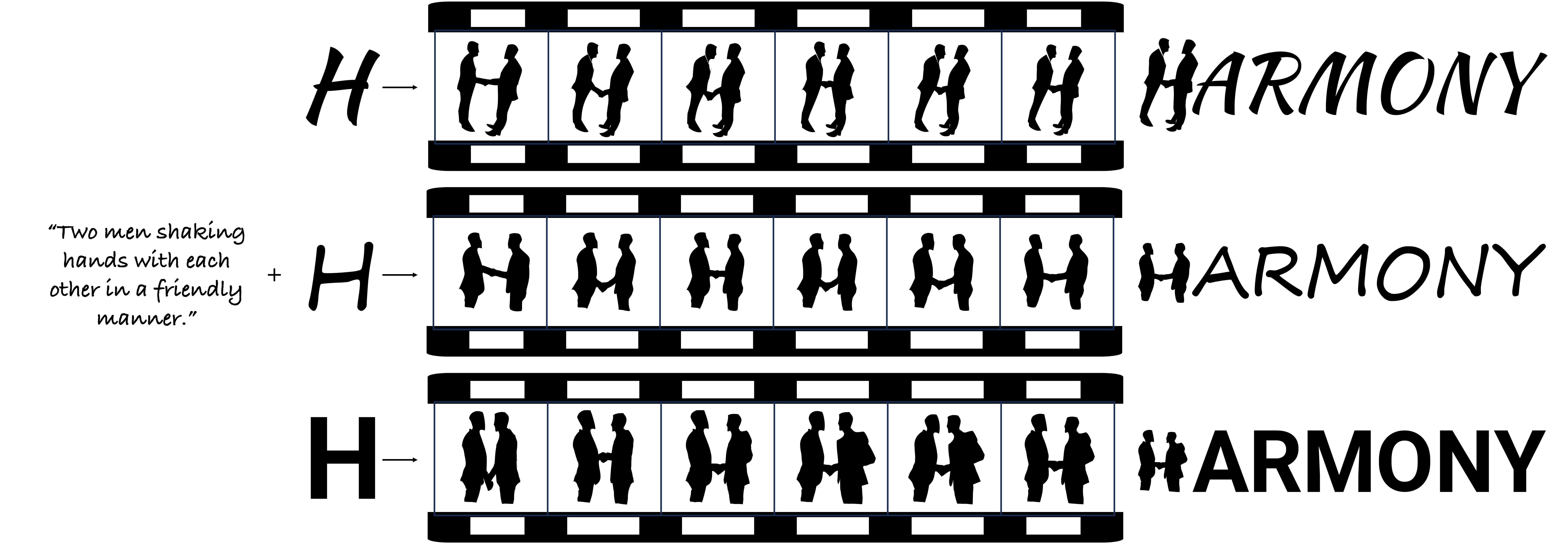}
\caption{Dynamic Typography over different fonts for the same animation sample. The corresponding fonts in the first, second, and third rows are KaushanScript-Regular, Segoe Print, and Roboto-Bold respectively. The animated letter ``H'' preserves the unique style of each font while faithfully depicting the prompts, allowing it to be seamlessly integrated into the word ``HARMONY'' under different fonts.}
\label{fig:gen_fonts}
\end{figure*} 

\begin{figure*}[!htb]
\centering
\includegraphics[width=1.0\textwidth]{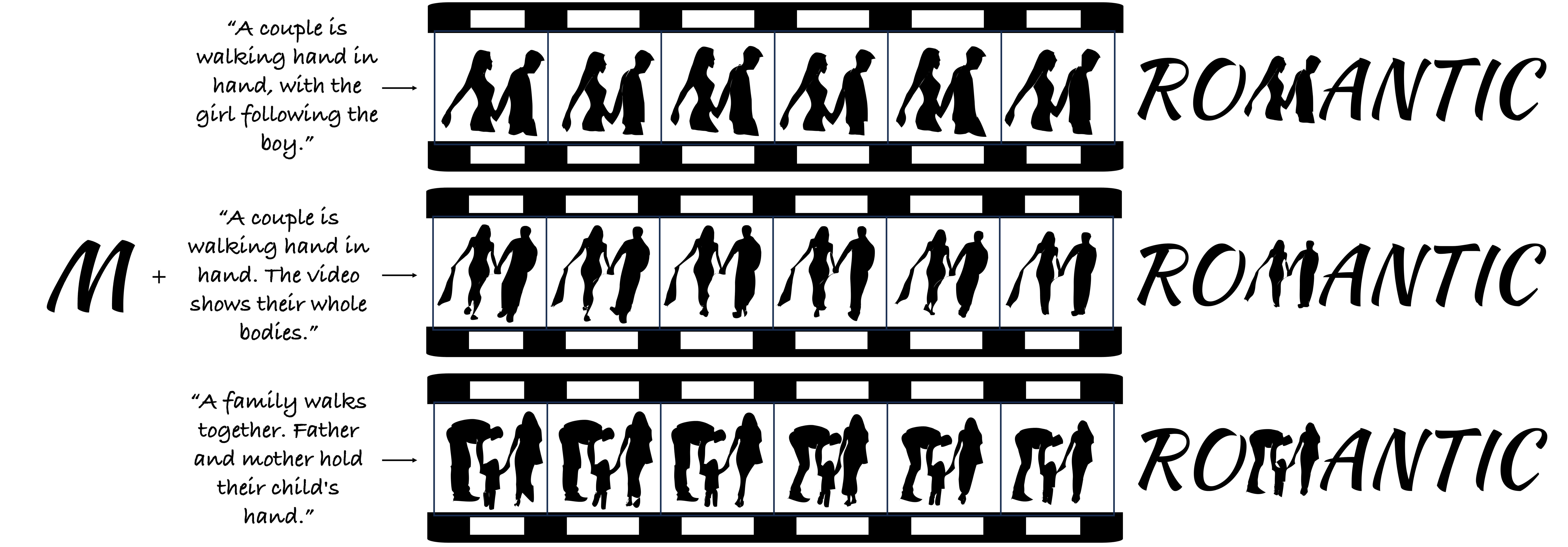}
\caption{Dynamic Typography over different prompts for the same letter ``M'' to be animated. Our approach displays different visual effects based on three different prompts, offering a completely different reading experience for the same word.}
\label{fig:gen_prompts}
\end{figure*}

\begin{CJK*}{UTF8}{gkai}
\begin{CJK*}{UTF8}{min}
\begin{figure*}[!htb]
\centering
\includegraphics[width=1.0\textwidth]{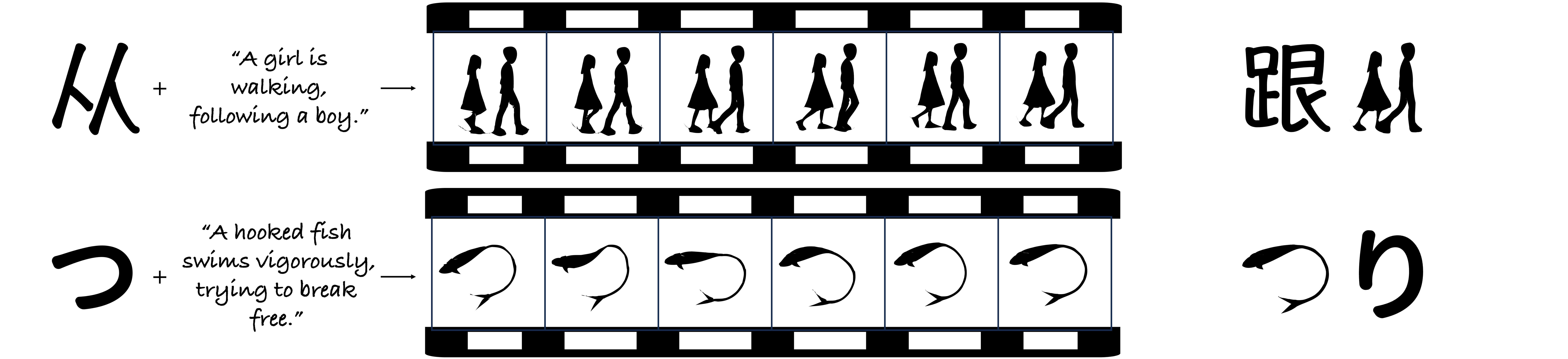}
\caption{Dynamic Typography over different languages. In the first row, we animate the Chinese character ``从'' in the word ``跟从'', meaning ``following'' in English. In the second row, we animate the hiragana character 「つ」 from the Japanese word 「つり」, meaning ``fishing'' in English. This demonstrates the potential of our proposed methodology to generate Dynamic Typography over different languages.}
\label{fig:gen_lan}
\end{figure*} 
\end{CJK*}
\end{CJK*}
\clearpage

\section{Generalizability over Different Fonts, Prompts and Languages}
Our method generalizes well in generating Dynamic Typography over different fonts, text prompts, and languages. 
When the letter to be animated takes different fonts, the generated animation preserves the unique style of each font, as illustrated in Fig. \ref{fig:gen_fonts}. 
In Fig. \ref{fig:gen_prompts}, by adjusting the input prompts, our method can generate animation with distinct visual effects for the same letter to be animated.
Fig. \ref{fig:gen_lan} demonstrates two Dynamic Typography samples in Chinese and Japanese, indicating the potential to generalize into different languages.

\section{User Study}
We conduct a user study to further compare our animation result with three baseline models: DynamiCrafter \cite{dynamicrafter}, Gen-2 \cite{gen-2}, and LiveSketch \cite{livesketch}, using the same dataset as the quantitative analysis. 
In the study, users are asked to select their preferred sample from four options, each produced by one of the four models. Their selections are guided by the three crucial requirements for Dynamic Typography as described in the method section in the main body, i.e., legibility preservation, semantic alignment, and temporal consistency. Users are required to answer 8 questions that are randomly sampled from the dataset.
We collect 62 responses in total. 
\begin{figure}[H]
    \centering
    \includegraphics[width=1\columnwidth]{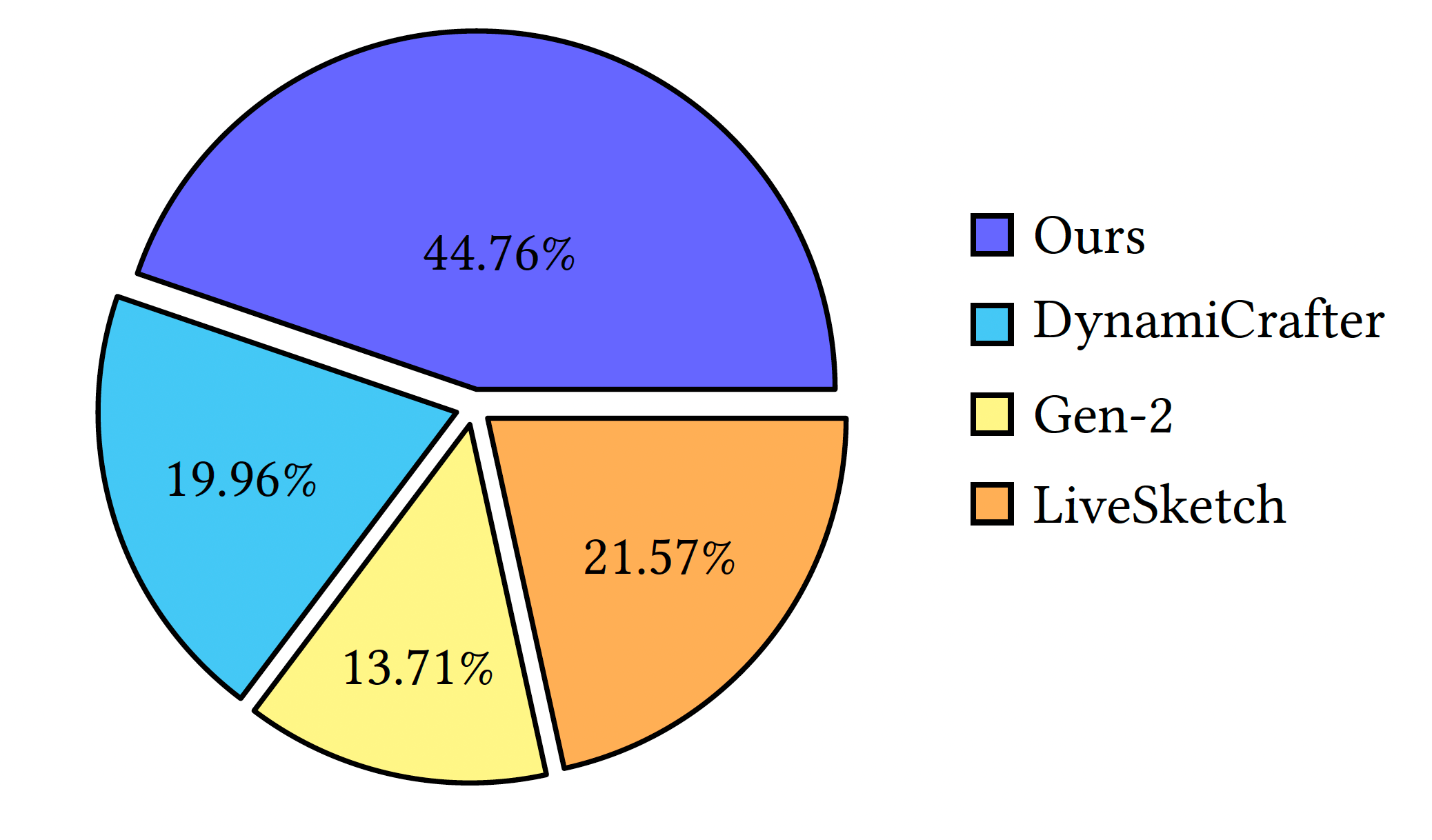}
    \caption{User Study results. Users are required to select the best animation based on legibility preservation, semantic alignment, and temporal consistency. The pie chart shows the percentage of each method being selected as the best animation.}
    \label{fig:user study} \vspace{-0.3cm}
\end{figure}
According to the final result shown in Fig. \ref{fig:user study}, our animation demonstrates transcendent distinction among the baseline models. Out of $44.76\%$ responses, the animation generated by our method is selected to be the best, indicating that our generated animations best meet the crucial requirements.

\section{Failure Case}\label{fail}
We observe that, in some samples, the semantic meaning and corresponding motion specified by the user-provided text prompt significantly diverged from the original shape of the letter.
In such cases, the model struggles to simultaneously maintain the shape of the letter and convey the vivid semantic information of the text prompt. 
As a result, the letter either undergoes minimal change in shape, retaining its original form, or it completely loses its original shape, compromising legibility.

\begin{figure}[H]
    \centering
    \includegraphics[width=1\columnwidth]{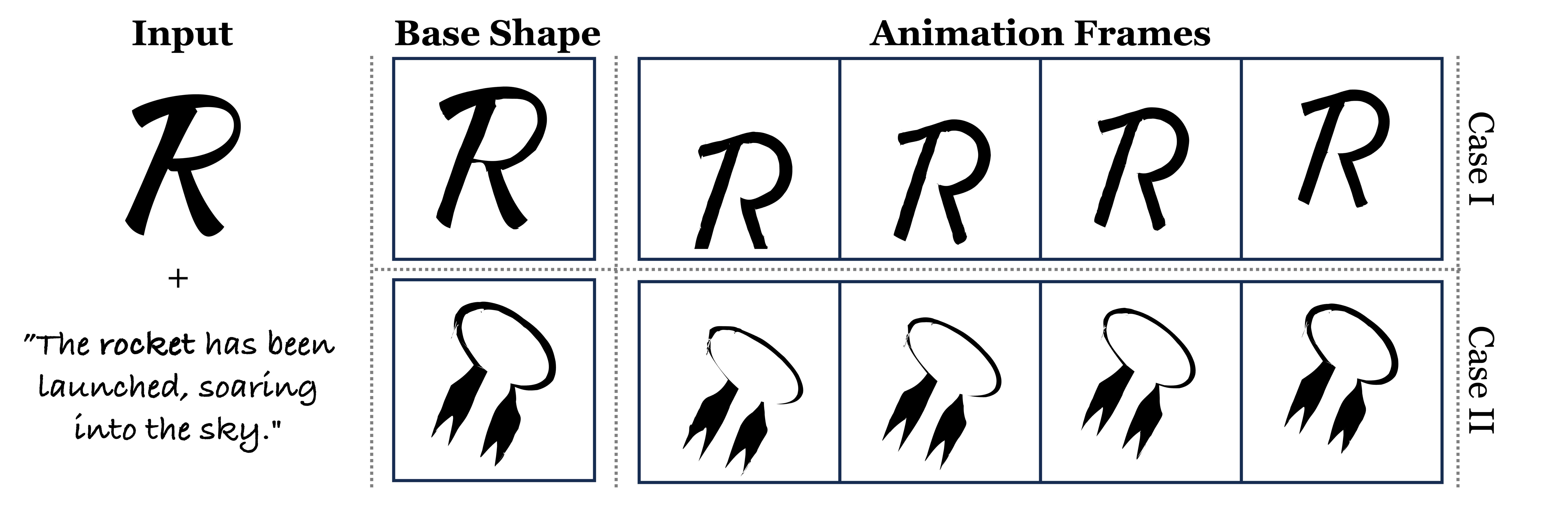}
    \caption{Failure case illustration. The first row is generated with the default weight for legibility and structure preservation loss, which suffers from minimal semantic deformation. In the second row, we reduce the weight of these regularization losses, which compromises the legibility.}
    \label{fig:failure} 
\end{figure}
For example, in Fig. \ref{fig:failure}, when we incorporate the legibility regularization, the shape of the letter ``R'' remains unchanged, maintaining its original form while performing the ``launch'' action. 
Conversely, when we reduce the weight of the legibility regularization, it transforms completely into the shape of a rocket, losing the characteristic contours of the letter  ``R'', thus sacrificing the legibility.

\section{GPT-4V as Dynamic Typography Designer}

As illustrated in Fig. \ref{fig:failure}, if the user-specified text prompt deviates too much from the chosen letter's shape, it can hinder the creation of vivid animations. We can utilize the powerful visual and semantic understanding capabilities of Vision Language Models (VLMs) to assist users in selecting appropriate letters and prompts.

In the experiment, we provide current state-of-the-art VLM, GPT-4V \cite{gpt4}, with a snapshot of an animation generated, along with the corresponding chosen word, letter, and text prompt as an example to facilitate in-context learning by GPT-4V. 
Subsequently, we request GPT-4V to design text animations by following the paradigm exemplified in the previous experiment. 
We require GPT-4V to generate outputs including the word, the selected letter, and the text prompt, explicitly demanding that it considers the similarity between the letter's original shape and its shape after deformation. 

\begin{figure*}[!htb]
\centering
\includegraphics[width=1.0\textwidth]{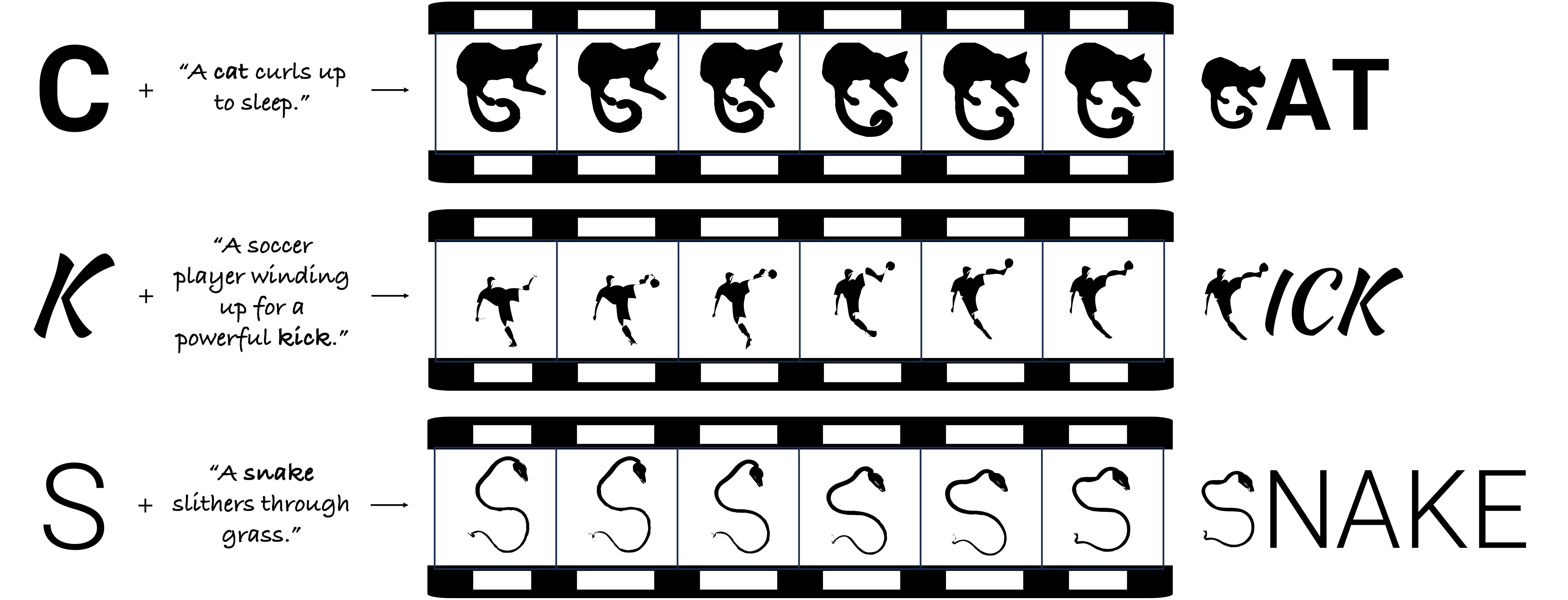}
\caption{
Some results generated by our method based on pairs of prompts and letters that are designed with the assistance of GPT-4V.}
\label{fig:GPT4}
\end{figure*} 

We list some samples designed by GPT-4V as follows:

\noindent
\begin{tabular}{>{\bfseries}l p{0.65\columnwidth}}
  Word:  & CAT\\
  Chosen Letter: & C\\
  Text Prompt: & ``A cat curls up to sleep.'' \\
  Animation Idea: & The ``C'' naturally curls tighter into a circular shape, resembling a cat curling up. \\
  & \\ 
  Word:  & KICK\\
  Chosen Letter: & K \\
  Text Prompt: & ``A soccer player kicks a ball.''\\
  Animation Idea: & The angled legs of the ``K'' mimic the motion of kicking, with one leg drawing back and then striking forward. \\
  & \\ 
  Word:  & SNAKE\\
  Chosen Letter:  & S\\
  Text Prompt: & ``A snake slithers through grass.'' \\
  Animation Idea: & The natural curve of ``S'' undulates slightly, resembling the slithering movement of a snake. \\
\end{tabular}

The generated animations designed by GPT-4V are shown in Fig. \ref{fig:GPT4}.
We found that GPT-4V has the potential to design proper pairs of prompts and letters that carefully consider the natural shapes of the letters and how they can effectively transform into the desired actions or characteristics with minimal deviation, ensuring the animations are feasible and visually coherent.

\clearpage

\end{document}